%% file: main.tex
\documentclass{article}
\usepackage{amssymb}

\usepackage[preprint]{corl_2025} 

\usepackage{float}
\usepackage{graphicx}  
\usepackage{caption}   
\usepackage{subcaption}

\usepackage{bold-extra}
\usepackage{enumitem}
\usepackage{booktabs}  
\usepackage{multirow}  
\usepackage{enumitem}
\usepackage{arydshln}
\usepackage{amsmath}
\usepackage{algorithm}
\usepackage{algpseudocode}
\usepackage{bm}
\usepackage{todonotes}
\usepackage{subcaption}
\usepackage{algorithm}
\usepackage{algpseudocode}
\usepackage{amsmath}

\usepackage{xcolor}    
\definecolor{lightred}{rgb}{0.988, 0.294, 0.0823}
\definecolor{lightblue}{rgb}{0.11, 0.541, 0.752}
\definecolor{lightgreen}{rgb}{0.1, 0.6, 0.1}

\title{\raisebox{-0.2\height}{\includegraphics[height=1.2em]{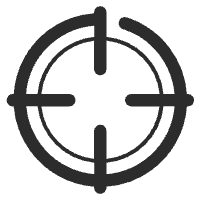}}\texttt{AimBot}: A Simple Auxiliary Visual Cue to Enhance Spatial Awareness of Visuomotor Policies}

\author{
  Yinpei Dai$^\dagger$$^*$, Jayjun Lee$^\ddagger$$^*$, Yichi Zhang$^\dagger$, Ziqiao Ma$^\dagger$, Jianing Yang$^\dagger$\\
  \textbf{Amir Zadeh$^\diamondsuit$, Chuan Li$^\diamondsuit$, Nima Fazeli$^{\dagger\ddagger}$$^\star$, Joyce Chai$^\dagger$$^\star$} \\
  $^\dagger$Computer Science and Engineering Department, University of Michigan\\
  $^\ddagger$Robotics Department, University of Michigan\\
 $^\diamondsuit$Lambda Labs\\
 $^*$ Equal Contribution $^\star$ Equal Advising\\
  \texttt{\{daiyp, jayjun, nfz, chaijy\}@umich.edu} \\
}

\newcommand{\papername}{\texttt{\textbf{AimBot}}}

\begin{document}
\maketitle


\input{sections/0_abstract}

\input{floating/teaser_figure}

\input{sections/1_introduction}
\input{sections/2_related_works}
\input{sections/3_methodology}

\input{sections/4_experiments}
\input{sections/5_conclusion}

\clearpage

\acknowledgments{This work was supported in part by NSF SES-2128623, NSF CAREER \#2337870, and NSF NRI \#2220876. We would like to thank Dr. Xinyi Wang for insightful discussion. We would also like to thank Lambda Labs for providing helpful GH200 computing resources.}


\input{sections/7_limitation}
\bibliography{example}  

\input{sections/6_appendix}

\end{document}

%% file: sections/0_abstract.tex
\begin{abstract}
In this paper, we propose \papername{}, a lightweight visual augmentation technique that provides explicit spatial cues to improve visuomotor policy learning in robotic manipulation.
\papername{} overlays shooting lines and scope reticles onto multi-view RGB images, offering auxiliary visual guidance that encodes the end-effector's state.
The overlays are computed from depth images, camera extrinsics, and the current end-effector pose, explicitly conveying spatial relationships between the gripper and objects in the scene.
\papername{} incurs minimal computational overhead (less than 1 ms) and requires no changes to model architectures, as it simply replaces original RGB images with augmented counterparts.
Despite its simplicity, our results show that \papername{} consistently improves the performance of various visuomotor policies in both simulation and real-world settings,
highlighting the benefits of spatially grounded visual feedback.
Codes and videos can be found at \url{https://aimbot-reticle.github.io/} 
\end{abstract}

%% file: floating/teaser_figure.tex

\begin{figure}[h]
    \centering
    \includegraphics[width=1.0\textwidth]{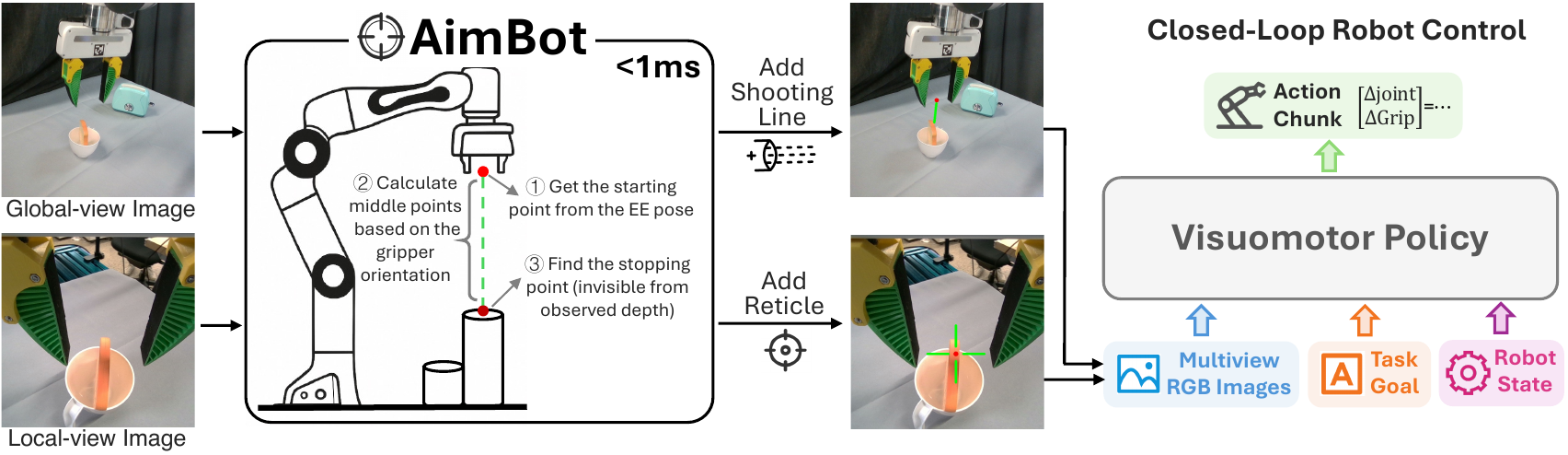}
    \caption{Overview of \papername{}, a lightweight visual guidance method that adds spatial cues onto RGB images for visuomotor policy learning. Given the robot’s end-effector pose, camera extrinsic and depth image, \papername{} computes shooting lines and reticle overlay to highlight the spatial relationship between the gripper and objects of interest. The augmented RGB images are used as input to visuomotor policies for closed-loop control, enhancing task performance with minimal overhead.}
    \label{fig:teaser}
    \vspace{-10pt}
\end{figure}

%% file: sections/1_introduction.tex
\section{Introduction}
\vspace{-6pt}
Robotic manipulation in unstructured environments demands 
visuomotor policies that can robustly and accurately predict continuous actions from raw RGB observations. Although recent advances, such as diffusion-based policies~\cite{diffusion-policy, khazatsky2024droid} and vision-language-action (VLA) models~\cite{pi0,open-vla,fast,hung2025nora, lin2025onetwovla}, have leveraged large-scale datasets to learn complex behaviors, they often lack explicit spatial grounding in visual input. As a result, these policies often exhibit limited spatial awareness of their end-effector (EE) pose and its relationship to surrounding objects \cite{qu2025spatialvla, trace-vla, li2025pointvla, song2025robospatial, zhou2025roborefer, firoozi2023foundation}.

Motivated by the intuitive visual feedback offered by scope reticles in optical sighting systems~\cite{jeung2019development,zhang2021augmented,opticsplanet_reticle_guide}, we present \texttt{\papername{}}, a lightweight and effective visual augmentation technique that enhances spatial awareness in robotic manipulation with visual targeting cues. \papername{} overlays auxiliary scope reticles and shooting lines onto multi-view RGB images by leveraging depth information, camera extrinsics, and the robot’s EE pose. These spatial cues project the intended grasp point and the orientation of the EE onto the image plane, offering an interpretable visualization of the alignment between the gripper and objects of interest (see Figure \ref{fig:teaser}). Beyond highlighting spatial relationships, these augmentations also intuitively encode the robot's proprioceptive state, making the EE position and orientation directly accessible in the visual domain.

\papername{} requires no changes to model architecture and introduces negligible computational overhead (less than 1 ms), while substantially enriching the spatial information available to any visuomotor policy. Through extensive experiments in both simulation and real-world environments, we demonstrate that \papername{} consistently improves overall task performance across diverse VLA backbones, with particularly strong gains on challenging, long-horizon tasks  that demand effective spatial alignment between the EE and objects.
We believe this simple yet effective approach provides a practical and scalable means to enhance learning-based robotic manipulation in 3D environments.


%% file: sections/2_related_works.tex
\section{Related Works}
\label{sec:related_works}

\subsection{Vision-Language-Action Models for Manipulation}
\vspace{-1mm}
Recent advances in pre-trained foundation models \cite{touvron2023llama,  bai2023qwen, he2022galaxy, achiam2023gpt} have catalyzed the development of vision-language-action (VLA) models, which significantly boost visuomotor policy learning in generalization, scalability, and robustness \cite{pi0, open-vla, xi-etal-2024-teaching}. OpenVLA~\cite{open-vla} and OpenVLA-OFT~\cite{open-vla-oft} extend the Llama model \cite{touvron2023llama} to large-scale real-world robot data, achieving strong performance and effective action generation. $\pi_0$~\cite{pi0}, $\pi_0$-FAST~\cite{fast} leverage the Gemma model \cite{gemma} and compression techniques to develop generalist policies capable of handling complex manipulation tasks. GR00T~\cite{bjorck2025gr00t} utilize a heterogeneous mixture of real-robot trajectories, human videos, and synthetic datasets for expressive humanoid control.
FuSe~\cite{jones2025beyond} moves beyond data scaling by leveraging multiple sensory modalities, demonstrating that natural language can universally ground vision, touch, and sound without requiring extensive multi-modal datasets. 
GO-1~\cite{bu2025agibot} improves long-horizon reasoning through a VQ-VAE latent action model trained on web-scale video data.
All these work highlight the rapid progress in robotic manipulation, providing a foundation upon which our work can build.

\subsection{Visual Augmentation/Guidance for Manipulation}
\vspace{-1mm}

With the success of visual prompting in VLMs~\cite{yang2023set,shtedritski2023does, dai2024racer, dai2024think, nasiriany2024pivot, li2024visual}, recent approaches have explored visual intermediaries to enhance generalization in robotic manipulation. 
RT-Affordance~\cite{rt-affordance} predicts affordance plans (i.e., key robot poses) conditioned on the visual input, enabling flexible learning across diverse supervision sources. 
RT-Trajectory~\cite{rt-trajectory} extends this idea by conditioning policies on coarse trajectory sketches, facilitating generalization to novel scenarios. 
GENIMA~\cite{genima} fine-tunes a diffusion model to overlay joint-action targets onto RGB images, which are then translated into joint positions. 
TraceVLA~\cite{trace-vla} introduces visual trace prompting that encodes spatiotemporal trajectories directly into visual inputs, enabling VLA models to better predict actions. RoboPoint~\cite{robopoint} leverages a VLM to predict keypoint affordances in terms of points on RGB images.
HAMSTER~\cite{hamster} introduces a hierarchical approach that separates high-level task planning from low-level motor control using intermediate 2D path prediction. 
However, all those methods require online model inference during deployment, significantly limiting their practicality for real-time control. In contrast, our work presents a lightweight visual augmentation technique that overlays interpretable 2.5D spatial cues onto RGB images, offering spatial information without incurring inference costs.  Moreover, by embedding EE state information directly into the pixel space, our method provides a novel and effective EE representation to enhance vision-language-action models.

%% file: sections/3_methodology.tex
\section{Methodology}

\subsection{Method Overview}

Visuomotor policies \cite{diffusion-policy,levine2016end} aim to directly map RGB camera observations and proprioceptive states to robot actions for sensorimotor control. Our goal is to enhance these policies with auxiliary visual cues, such as shooting lines and crosshair reticles, to improve spatial alignment and task success. The proposed technique, \papername{}, is model-agnostic and requires no modifications to underlying policy architectures. It operates by augmenting the original multi-view RGB images with visual guidance, embedding spatial information directly into the pixel space. Fine-tuning is then performed exclusively on these new images, enabling any visuomotor policy or vision-language-action model to leverage the enhanced spatial cues.

\subsection{\papername{} Visual Guidance}
\label{method:aimbot}

Suppose we have the camera extrinsics $\bold E \in \mathbb{R}^{4 \times 4}$ and intrinsics $\bold K \in \mathbb{R}^{3 \times 3}$, along with the RGB image $I$ and depth image $D$  captured at the current timestep from  camera $c$. 
Given a 3D point $\mathbf{p}_{\text{wld}} \in \mathbb{R}^{3}$ in the world frame, we can project it into camera frame with pinhole camera model \cite{forsyth2002computer}:
\begin{equation}
\begin{bmatrix}
\mathbf{p}_{\text{cam}} \\
1
\end{bmatrix}
= \bold E \cdot 
\begin{bmatrix}
\mathbf{p}_{\text{wld}} \\
1
\end{bmatrix}, \quad \text{where} \; \mathbf{p}_{\text{cam}} = (x_c, y_c, z_c)^\top.
\end{equation}
Then, we can continue to project $\mathbf{p}_{\text{cam}}$ to 2D image coordinates  $(u_c, v_c)$ using the intrinsic matrix:
\begin{equation}
\begin{bmatrix}
u_c \\
v_c \\
1
\end{bmatrix}
\propto
\bold K 
\begin{bmatrix}
x_c / z_c \\
y_c / z_c \\
1
\end{bmatrix}.
\end{equation}
We define a point as \textit{visible} if the projected pixel lies within the image bounds and the projected depth is smaller than the observed depth (i.e, not being blocked by any objects). Formally, let $(u_c, v_c)$ be the projected pixel location and $z_c$ the projected depth. The point is visible if:
\begin{equation}
0 \leq u_c < W, \quad 0 \leq v_c < H, \quad \text{and} \quad z_c + \epsilon < D[v_c, u_c]
\end{equation}
where $H$ and $W$ are the image height and width, respectively, $D[v_c, u_c]$ is the observed depth at pixel $(u_c, v_c)$, and $\epsilon>0$ is a small threshold. 
Next, we describe how to determine the starting and stopping point of a forward line based on visible points and how to add \papername{} visual guidance to RGB images. The detailed algorithm can be found in Appendix \ref{sec: appx-alg}.

\paragraph{Starting Point.} 
We always set the origin of the gripper frame attached to the end-effector (EE) as the starting point,
which is denoted as $\mathbf{p}^{ee}_{\text{wld}}$ in the world frame.
The pixel location of the starting point in the image is denoted as $(u^{ee}_c, v^{ee}_c)$ for camera $c$. The camera can be either fixed or movable.

\paragraph{Stopping Point.}
Starting from $\mathbf{p}^{ee}_{\text{wld}}$, we iteratively move forward along the direction vector $\mathbf{d} \in \mathbb{R}^3$, derived from the EE's orientation (e.g., the $z$-axis of the gripper frame):
\[
\mathbf{p}_{\text{wld}}^{(i+1)} = \mathbf{p}_{\text{wld}}^{(i)} + \delta \cdot \mathbf{d}, \quad \text{with } \mathbf{p}_{\text{wld}}^{(0)} = \mathbf{p}^{ee}_{\text{wld}}.
\]
where $ \delta>0$ is a small value denoting step length.
At each step, we project $\mathbf{p}_{\text{wld}}^{(i)}$ to image coordinates and check visibility using the same procedure as above. The iteration stops when a certain tolerance number of invisible $\mathbf{p}_{\text{wld}}^{(i)}$ or the maximum step is reached. Then we choose the last  $\mathbf{p}_{\text{wld}}^{(L)}$ as our stopping point, where $L$ is the total iteration number. We denote the pixel location of the stopping point in the camera image as $(u^{sp}_c, v^{sp}_c)$. The total projection distance from $\mathbf{p}_{\text{wld}}^{ee}$ to $\mathbf{p}_{\text{wld}}^{(L)}$ is $\delta L|\mathbf{d}|$.

\paragraph{Augment Global-View with Shooting Lines.} In robotic manipulation, global-view observations are typically captured from static external cameras, such as front-facing or shoulder-mounted cameras. For these views, we overlay a shooting line on the image, extending from the pixel location corresponding to the starting point $(u^{ee}_c, v^{ee}_c)$ to the projected stopping point $(u^{sp}_c, v^{sp}_c)$. This line serves as an explicit visual indicator of the EE's position and orientation. 
In our default implementation, we also use color to convey gripper state: a \textcolor{lightgreen}{green} line with a \textcolor{red}{red} starting point indicates that the gripper is open, while a \textcolor{purple}{purple} line with a \textcolor{blue}{blue} starting point indicates closed gripper.

\vspace{-6pt}
\paragraph{Augment Local-View with Reticles.} Wrist-mounted cameras are commonly employed to capture dynamic, close-range egocentric observations that provide detailed local views of the scene. To convey the EE's state, we overlay scope reticles onto the wrist-view image. Specifically, we render a crosshair-style reticle centered at the projected stopping point, indicating the direction in which the gripper is pointing. 
Note that, the pixel location $(u^{sp}_{\text{wrist}}, v^{sp}_{\text{wrist}})$ varies depending on the projection distance $\delta L|\mathbf{d}|$, which reflects the distance from the EE to the nearest surface. 
When the projection distance is large (e.g., the gripper is far from the table), $(u^{sp}_{\text{wrist}}, v^{sp}_{\text{wrist}})$  appears closer to the center of the wrist image due to perspective effects. Conversely, when the projection distance is small (e.g., the gripper is close to an object that obstructs the projection path), $(u^{sp}_{\text{wrist}}, v^{sp}_{\text{wrist}})$ aligns more closely with the center of the gripper pads in the image. 

To encode additional information about spatial proximity,  
the length of the reticle lines is also modulated based on the projection distance: the reticle lines are shorter when the distance is large and longer when the distance is small, providing a visual indication of the estimated distance to the nearest orthogonal surface (see Appendix \ref{sec: appx-alg} for details). This augmentation provides a visual cue that helps the policy develop a good understanding of spatial depth and distance in the local environment. In our default implementation, the crosshair lines are rendered in \textcolor{lightgreen}{green}, and the center point is colored \textcolor{red}{red} or \textcolor{blue}{blue} to indicate whether the gripper is currently open or closed, respectively.

%% file: sections/4_experiments.tex
\vspace{-4pt}
\section{Experiments}
\vspace{-4pt}
    

We choose three latest vision-language-action models, $\pi_0$\cite{pi0}, $\pi_0$-FAST \cite{fast} and OpenVLA-OFT~\cite{open-vla-oft}, as our visuomotor policy backbones, and conduct experiments on both simulated and real environments to evaluate our approach.  For the simulation study, we choose the LIBERO~\cite{libero} benchmark as the testbed. For the real-world study, we design five challenging tasks to test our approach. 

\vspace{-2pt}
\subsection{Simulation Experiments}
\vspace{-2pt}

\begin{figure}[t]
    \centering
\includegraphics[width=0.98\linewidth]{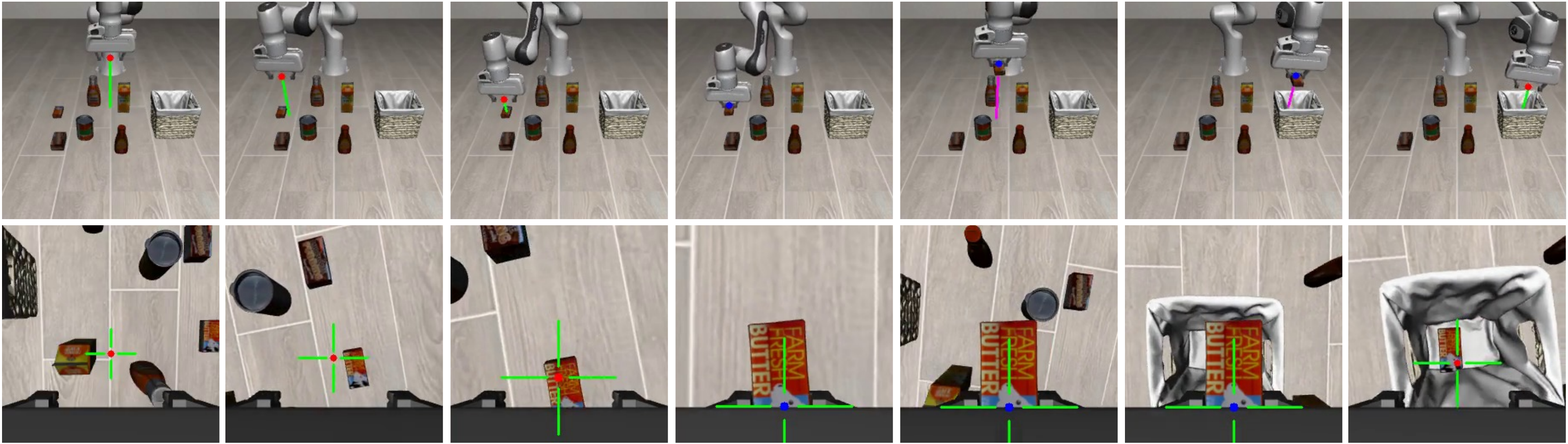}
    \caption{An example of \papername{}-augmented LIBERO observations. We add a shooting line and a crosshair reticle for the front-view (top row) and wrist-view (bottom row) cameras, respectively.}
    \label{fig:libero-example}
    \vspace{-14pt}
\end{figure}

\noindent\textbf{Simulation Setup:} The LIBERO benchmark consists of four task suites (LIBERO-Spatial, LIBERO-Object, LIBERO-Goal, and LIBERO-Long) designed for studying lifelong learning in robotic manipulation. LIBERO-Spatial varies scene layouts with the same objects to test spatial reasoning, LIBERO-Object varies objects within fixed layouts to test object understanding, and LIBERO-Goal varies task goals while keeping objects and layouts fixed to test task-oriented behavior. LIBERO-Long includes long-horizon tasks with diverse objects, layouts, and goals, and is the most challenging suite as it demands more accurate gripper-object alignment over extended trajectories.
To generate training data, we regenerate all the demonstrations following \cite{open-vla}, and augment both multi-view RGB images with \papername{}  at a resolution of 256$\times$256 pixels (shown in Figure~\ref{fig:libero-example}). 

\noindent\textbf{Implementation Details:} Following \cite{pi0}, we train a single multi-task policy for all tasks rather than different policies for each task. All models are optimized using AdamW \cite{loshchilov2017decoupled} for 30k steps with a batch size of 32. For $\pi_0$ and $\pi_0$-FAST, we adopt a learning rate of 2e-4 with 1k warm-up steps and cosine decay. For OpenVLA-OFT, we use a fixed learning rate of 5e-4 and an L1 regression loss for action training, along with LoRA \cite{hu2022lora} optimization (rank 32, alpha 16).
The action horizon is set to 50 steps for $\pi_0$, and 10 steps for the other two, following their default settings. Observations consist of front-view and wrist-view RGB images, along with proprioceptive states provided by the simulator.
All models operate in a delta Cartesian action space (6 dimensions) comprising changes in EE position and EE orientation represented in Euler angles. An additional dimension is used to represent the gripper open/close action.
During evaluation, we execute 5 steps per action prediction for $\pi_0$ and $\pi_0$-FAST, and 8 steps for OpenVLA-OFT, enabling closed-loop control.


\paragraph{Experimental Results.}
For baselines trained without \papername{}, we re-implement all models on the original LIBERO benchmarks and report the higher results between our re-implementations and their released numbers. The overall results are summarized in Table~\ref{tab:libero}. Across all three backbone models, integrating and finetuning with \papername{} consistently improves or matches baseline performance in the majority of cases, with particularly large gains on the hardest LIBERO-Long tasks. For instance, $\pi_0$-FAST improves from 81.6 to 87.1 and $\pi_0$ improves from 85.2 to 91.0  with \papername{}. OpenVLA-OFT also achieves a notable improvement from 87.5 to 91.2 on LIBERO-Long.
This indicates that the spatially grounded visual guidance provided by \papername{} can enhance long-horizon manipulation. 
For other task suites, where baseline performances are already relatively high, \papername{} leads to smaller improvements. Nevertheless, the average success rates still increase: +1.2 for OpenVLA-OFT, +1.6 for $\pi_0$-FAST, and +1.7 for $\pi_0$. Overall, \papername{} provides the greatest benefits in more challenging, long-horizon settings while offering modest gains in easier tasks.
To further study the impact of reticle design choices, we compare several \papername{} variants in Appendix~\ref{sec: appx-ablation} and adopt the best-performing setting (same setting as shown in Figure~\ref{fig:libero-example}) for real-world experiments.

\begin{table}[t]
\centering
\scalebox{0.9}{
\begin{tabular}{lccccc}
\toprule
\textbf{Model} & 
\begin{tabular}[c]{@{}c@{}} \texttt{LIBERO} \\ \texttt{Spatial} \end{tabular}
& \begin{tabular}[c]{@{}c@{}} \texttt{LIBERO} \\ \texttt{Object} \end{tabular} & \begin{tabular}[c]{@{}c@{}} \texttt{LIBERO} \\ \texttt{Goal} \end{tabular} & \begin{tabular}[c]{@{}c@{}} \texttt{LIBERO} \\ \texttt{Long} \end{tabular} & \begin{tabular}[c]{@{}c@{}} \textsc{Average} \\ \textsc{Success Rate} \end{tabular} \\
\midrule
OpenVLA-OFT \cite{open-vla-oft} & \textbf{96.2} & 97.3 & 93.9 & 87.5 & 93.8 \\
OpenVLA-OFT + \papername{} & 95.2 \textcolor{red}{(–1.0)} & \textbf{99.1} \textcolor{lightgreen}{(+1.8)} & \textbf{94.2 }\textcolor{lightgreen}{(+0.3)} & \textbf{91.2} \textcolor{lightgreen}{(+3.7)} & \textbf{95.0} \textcolor{lightgreen}{(+1.2)} \\
\midrule
$\pi_0$-FAST \cite{fast} & 96.5 & \textbf{96.8} & 93.6 & 81.6 &  92.1\\
$\pi_0$-FAST + \papername{} & \textbf{96.9} \textcolor{lightgreen}{(+0.4)} & \textbf{96.8} \textcolor{lightgreen}{(+0.0)} & \textbf{94.0} \textcolor{lightgreen}{(+0.4)} & \textbf{87.1} \textcolor{lightgreen}{(+5.5)} & \textbf{93.7} \textcolor{lightgreen}{(+1.6)}\\
\midrule
$\pi_0$ \cite{pi0} & 96.8 & \textbf{98.8} & 95.8 & 85.2 & 94.2\\
$\pi_0$ + \papername{} & \textbf{96.9} \textcolor{lightgreen}{(+0.1)} & 98.4 \textcolor{red}{(–0.4)} & \textbf{97.2} \textcolor{lightgreen}{(+1.4)} & \textbf{91.0} \textcolor{lightgreen}{(+5.8)} & \textbf{95.9} \textcolor{lightgreen}{(+1.7)} \\
\bottomrule
\end{tabular}
}
\vspace{5pt}
\caption{Performance comparison on the LIBERO simulation benchmark. \textcolor{lightgreen}{Green} and \textcolor{red}{Red} numbers indicate performance gains and losses, respectively. Each task suite averages over four runs.}
\vspace{-10pt}
\label{tab:libero}
\end{table}

\input{tables/main_results_table_v1}

\subsection{Real World Experiments}

\begin{figure}[t]
    \centering
\includegraphics[width=0.98\linewidth]{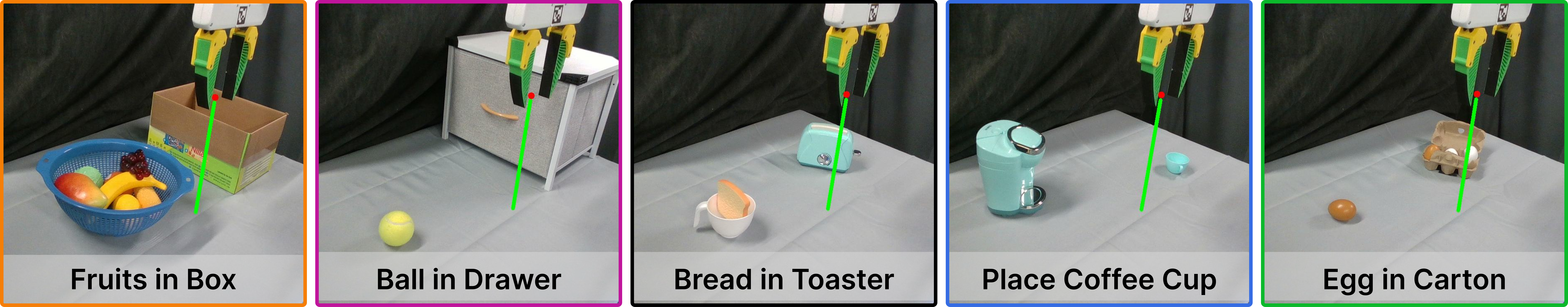}
    \caption{We design five contact-rich, long-horizon real-world tasks for policy evaluation.}
    \vspace{-15pt}
    \label{fig:real-task}
\end{figure}

\paragraph{Real World Setup.}
We conduct our experiments using a 7-DoF Franka Emika Panda robot equipped with a pair of UMI fin-ray fingers~\cite{umi-gripper} mounted on the default Franka hand gripper. The robot operates in a tabletop environment with three RGB-D cameras similar to the DROID setting \cite{khazatsky2024droid}: two Intel RealSense D435 cameras each positioned on the left and right shoulder, and one Intel RealSense D405 camera mounted on the wrist. The shoulder cameras are calibrated prior to experimentation to obtain accurate camera extrinsics, while the extrinsics of the wrist-mounted camera are dynamically computed based on the end-effector (EE) pose. For the depth images, we use the built-in  Intel RealSense SDK to preprocess and align the depth images to the RGB images.

\noindent{\bf Task Design and Data Collection.}
We designed five challenging tasks for the real world experiment: (1) \textbf{\texttt{Fruits in Box}}, where the robot is tasked to move all fruits inside a basket into a box; (2)  \textbf{\texttt{Tennis Ball in Drawer}}, where the robot needs to open a drawer, place a tennis ball inside, and close the drawer; (3) \textbf{\texttt{Bread in Toaster}}, where the robot has to grasp a bread and insert into a narrow toaster slot; (4) \textbf{\texttt{Place Coffee Cup}}, where the robot must grab a coffee cup by its handle, reorient it, and place it onto a coffee machine; and (5) \textbf{\texttt{Egg in Carton}}, where the robot needs to pick up one or more eggs, place them into an egg carton, and firmly close the carton lid. 
These tasks vary in length, complexity, and contact-richness, and require a combination of prehensile (e.g., pick-and-place) and non-prehensile (e.g., pulling a drawer handle, closing a carton lid) skills. Figure~\ref{fig:real-task} illustrates the table settings, and Appendix~\ref{appendix:dataset} details the policy rollout as sequences of RGB images augmented by \papername{} for each task.
For the real-world data collection, we teleoperate the robot using an Oculus Quest 2 device, recording demonstrations at 15 Hz. We collect 80–150 demonstrations per task, resulting in a total of 548 episodes with 166k timesteps of data samples.

\noindent{\bf Implementation Details.}
Following the simulation study, we fine-tune $\pi_0$, $\pi_0$-FAST, and OpenVLA-OFT to evaluate the effectiveness of \papername{} in real-world settings. We compare models trained on raw RGB images from multiple camera streams with models trained on RGB images augmented with \papername{}. All models are optimized for 50k steps with a batch size of 32. We use a learning rate of 1e-4 for $\pi_0$ and $\pi_0$-FAST, which we found to perform better, and set the action horizon to 10 steps for all models. Other hyperparameters are kept consistent with the simulation experiments. Observations include left-shoulder, right-shoulder, and wrist RGB images, along with the robot’s proprioceptive state. In terms of action space, $\pi_0$ and $\pi_0$-FAST predict delta joint angles (7 dimensions), while OpenVLA-OFT predicts delta Cartesian actions (6 dimensions); all models additionally output a binary gripper open/close command as an extra dimension. During online execution, we found that executing 8 steps per action prediction for $\pi_0$ and $\pi_0$-FAST, and 2 steps for OpenVLA-OFT, yields better performance.

\begin{figure}[t]
    \centering
    \begin{minipage}{0.48\textwidth}
        \centering
        \includegraphics[width=\textwidth, keepaspectratio]{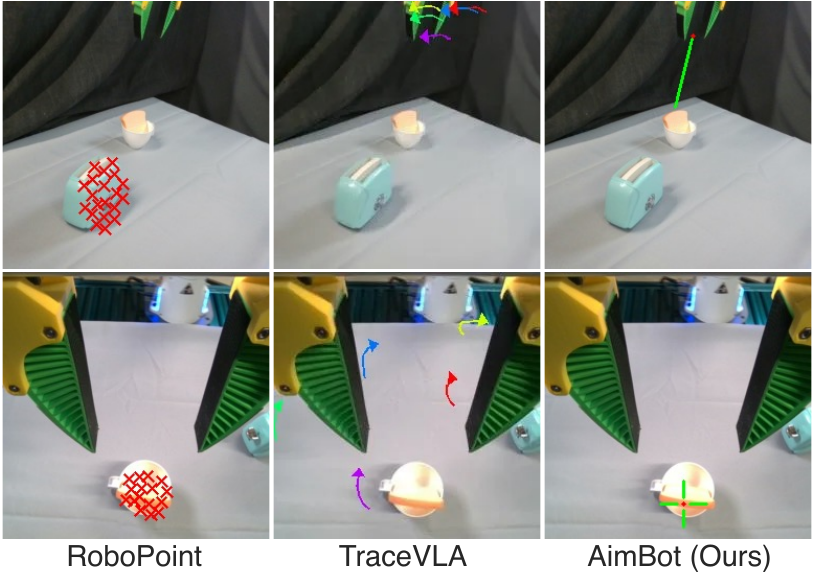}
        \caption{Comparison of different visual guidance methods.}
        \vspace{-15pt}
        \label{fig:baseline_visual_aug}
    \end{minipage}
    \hfill
    \begin{minipage}{0.48\textwidth}
        \centering
        \includegraphics[width=\textwidth, keepaspectratio]{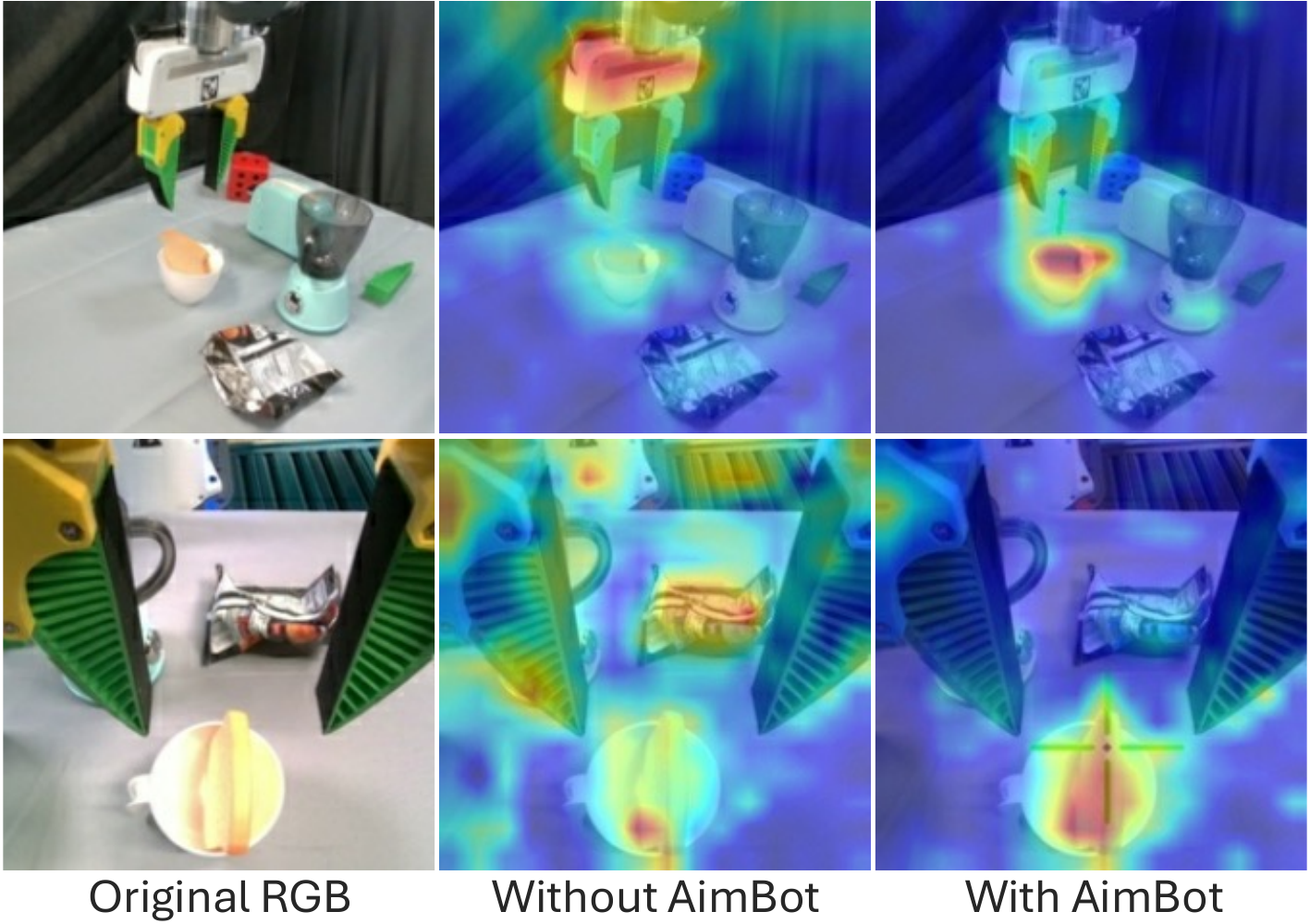}
        \caption{Visualization of attention weights trained with and without \papername{}.}
        \vspace{-15pt}
        \label{fig:attention}
    \end{minipage}
\end{figure}

\textbf{Baselines.} To compare \papername{} with other visual guidance methods, we evaluate two open-sourced baselines: RoboPoint \cite{robopoint} and TraceVLA \cite{trace-vla}, by training $\pi_0$ on images augmented using their respective strategies. RoboPoint predicts spatial affordances as 2D pixel coordinates on semantically relevant regions of the image, while TraceVLA visualizes temporal motion history through colored arrows (i.e., "traces"), offering cues about past trajectories. 
For RoboPoint, we query the robopoint-v1-vicuna-v1.5-13b checkpoint, using the prompt \textit{``The task is} \texttt{{<task\_goal>}}. \textit{Find relevant points on the image to perform the task."}, where \texttt{{<task\_goal>}} corresponds to the language description of each real-world task. 
We apply their default visualization tool to overlay red crosses at the predicted affordance points onto the RGB images as the visual augmentation for policy training. For TraceVLA, we apply their co-tracker model to track objects across three camera views and overlay the resulting traces onto the images. Example comparisons of the visual augmentations are shown in Figure~\ref{fig:baseline_visual_aug}.
In addition, we compare against $\pi_0$ trained with both raw RGB images and depth images to assess the effect of directly incorporating depth information. Concretely, we convert the depth image in grayscale RGB and treat it as three more images (a total of six images) to train $\pi_0$. 




\noindent{\bf Experimental Results.}
Table~\ref{tab:real-world} reports task success numbers (out of 10 trials per task) across five real-world manipulation tasks using different policy and visual guidance configurations. Overall, \papername{} significantly improves performance across all models and tasks. For example, OpenVLA-OFT’s total success increases from 21 to 36 successful trials when augmented with \papername{}, with significant gains in more challenging tasks such as \textbf{\texttt{Place Coffee Cup}} and  \textbf{\texttt{Bread in Toaster}}. Similarly, $\pi_0$ improves from 27 to 43 successful trials, while $\pi_0$-FAST achieves the highest overall performance with 47 successful trials when combined with \papername{}, outperforming all other models.

In contrast, alternative visual guidance baselines underperform: $\pi_0$ trained with RoboPoint and TraceVLA reaches only 27 and 25 successful trials, respectively, similar to the $\pi_0$ trained without any guidance. Unlike these methods, \papername{} provides direct spatial targeting by encoding the end-effector (EE) state as shooting lines and reticles, offering richer and clearer spatial cues without occluding objects or omitting gripper state information. Additionally, both TraceVLA and RoboPoint require online model inference, introducing significant computational overhead that limits their practicality for real-time robot control. On average, TraceVLA requires approximately 0.3 seconds to process a single image, while RoboPoint takes over 5 seconds. In contrast, \papername{} is highly efficient, requiring less than 1 ms per image, making it suitable for real-world deployment.

While using raw depth images as additional visual input yields a modest improvement over RGB-only inputs  (increasing success from 27 to 32), it still falls short of \papername{}’s performance (43), likely due to noise and inconsistency in real-world depth sensing. Despite extensive preprocessing, real-world depth data remains highly noisy and unreliable. In contrast, \papername{} is inherently robust and less redundant, as it only compares the projected point depth with the camera depth to determine visibility for visual cue generation, without relying on complete or very accurate depth information.

\vspace{-5pt}
\subsection{Further Analysis}
\vspace{-5pt}

To better understand how \papername{}'s visual guidance influences the model's internal representations, spatial reasoning, and generalization to unseen scenarios, we investigate the following questions:

\vspace{-5pt}
\paragraph{Does \papername{} enhance the robot's spatial awareness for object alignment?}
To address this, we first examine how training with \papername{} shapes the model's attention toward task-relevant objects. Specifically, we feed the same input image into the OpenVLA-OFT language model backbone and extract the attention weights from Layer 1, Head 11. We then compute the summed attention scores from the action head to the input RGB image patches and upsample them to the original image resolution to generate the heatmaps shown in Figure~\ref{fig:attention}. As illustrated, models trained with \papername{} exhibit more concentrated attention on the task-relevant objects, which facilitates better task understanding and execution. In contrast, the baseline model without \papername{} shows dispersed attention across the entire scene, making it harder for the robot to align accurately with target objects. More visualization examples can be found in Appendix \ref{appx: attention}.

To further investigate \papername{}'s impact on spatial alignment, we conduct a failure and error analysis based on real-world robot experiments. We categorize all failure trials into different misalignment types: grasping position misalignment, grasping orientation misalignment, placing position misalignment, placing orientation misalignment, and other non-misalignment failures (e.g., getting stuck, failed non-prehensile skills, or performing actions in a wrong task order). Examples of failures are illustrated in Appendix \ref{sec: appx-failure}. The aggregated results for three VLA models are summarized in Table~\ref{tab:failures_type}.
As shown, models trained with \papername{} demonstrate a substantial reduction in misalignment-related failures, particularly in grasping. This indicates that \papername{} effectively enhances the model's spatial understanding and improves alignment capability during manipulation.

\begin{table}[t]
\centering
\begin{minipage}{0.48\textwidth}{
\centering
\scalebox{0.85}{
\begin{tabular}{lcc}
\toprule
\textbf{Misalignment Type} & \textbf{w/o \papername{}} & \textbf{w/ \papername{}} \\
\midrule
Grasping Position & 22 & 7 \\
Grasping Orientation & 6 & 0 \\
Placing Position & 18 & 7 \\
Placing Orientation & 3 & 3 \\
Other Failures & 11 & 7 \\ 
\bottomrule
\end{tabular}}
}
\vspace{3pt}
\caption{Total failure counts across different misalignment types with and without \papername{}.}
\label{tab:failures_type}
\end{minipage}%
\hspace{0.02\textwidth}
\begin{minipage}{0.45\textwidth}{
\scalebox{0.85}{
\centering
\begin{tabular}{lc}
\toprule
\textbf{Models} & \texttt{LIEBRO-Long} \\
\midrule
$\pi_0$ $+$ \papername{} & 91.0 \\
$\pi_0$ $+$ \papername{} $-$ proprio. & 88.0 \\
$\pi_0$ & 85.2 \\
$\pi_0$ $-$ proprio. & 83.2 \\
$\pi_0$ $+$ \papername{} (random) & 77.4 \\ 
\bottomrule
\end{tabular}}
}
\vspace{3pt}
\caption{Ablation of $\pi_0$ variants with or without \papername{} and proprioceptive states.}
\label{tab:pi0_variant}
\end{minipage}
\vspace{-20pt}
\end{table}

\paragraph{Does \papername{} provide a better representation than proprioceptive state encoding?}
To investigate this question, we conduct an ablation study using $\pi_0$ on the LIBERO benchmark. The original $\pi_0$ setup includes a proprioceptive state vector encoding the end-effector state. We introduce two ablated variants: (1) $\pi_0$ $+$ \papername{} $-$ proprio., where \papername{} is incorporated while the proprioceptive input is replaced with a zero vector; and (2) $\pi_0$ $-$ proprio., where neither \papername{} nor proprioceptive input is used, with the proprioceptive state similarly zeroed out. Table~\ref{tab:pi0_variant} reports the performance on the LIBERO-Long task suite (more results are given in Appendix~\ref{sec: appx-prior}).
As shown, removing proprioceptive input reduces task success to 83.2\%, compared to 85.2\% when using proprioception alone. Incorporating \papername{} without proprioception improves performance to 88.0\%, and combining both leads to the highest task success of 91.0\%, suggesting that \papername{} provides a strong alternative representation and offers complementary benefits to standard proprioceptive state encoding.

To further validate that \papername{} effectively encodes useful spatial priors into the visual input, we introduce an additional variant, $\pi_0$ + \papername{} (random), where the auxiliary visual cues (shooting lines and reticles) are deliberately randomized by adding noise and misaligned from the true end-effector pose during both training and testing. As shown in Table~\ref{tab:pi0_variant}, this model achieves a significantly lower success rate of 77.4\%, compared to 91.0\% with correctly aligned \papername{}. This performance gap indicates that the spatial visual guidance provided by \papername{} is not only interpretable but also meaningfully enhances the model's spatial awareness and downstream policy performance.

\paragraph{Can \papername{} improve generalization to out-of-distribution (OOD) scenarios?}
We evaluate the ability of \papername{} to enhance OOD generalization in real-world tasks by introducing several distribution shifts at test time, including changes in object and receptacle heights, variations in table background colors, and lighting conditions that differ significantly from the training environment (examples can be found in Appendix \ref{sec: appx-ood}). For each task, we conduct 3 trials, resulting in a total of 15 evaluation episodes using the $\pi_0$-FAST model. In this  experiment, our method achieves \textbf{12} successful trials, whereas the baseline model achieves only 7. This suggests that \papername{} provides robust visual cues that remain effective under distribution shifts, allowing the model to leverage reliable spatial guidance even in unseen environments and thus improving generalization performance.




%% file: tables/main_results_table_v1.tex
\begin{table*}[t]
\centering
\scalebox{0.88}{
\begin{tabular}{lcccccccc}
\specialrule{0.3mm}{0mm}{0mm}
\textbf{Model} & \begin{tabular}[c]{@{}c@{}} \texttt{Fruits} \\ \texttt{in Box} \end{tabular}  & \begin{tabular}[c]{@{}c@{}} \texttt{Tennis Ball} \\ \texttt{in Drawer} \end{tabular} & \begin{tabular}[c]{@{}c@{}} \texttt{Bread in} \\ \texttt{Toaster} \end{tabular} & \begin{tabular}[c]{@{}c@{}} \texttt{Place} \\ \texttt{Coffee Cup} \end{tabular} & \begin{tabular}[c]{@{}c@{}} \texttt{Egg in} \\ \texttt{Carton} \end{tabular} & \begin{tabular}[c]{@{}c@{}} \textsc{Total} \\ \textsc{Success} \end{tabular} \\ 
\toprule
OpenVLA-OFT & 7/10 &  6/10 & 4/10 & 2/10 & 2/10  & 21/50\\
\addlinespace[3pt]
OpenVLA-OFT + \papername{}  & \textbf{9/10}   & \textbf{7/10} & \textbf{9/10} & \textbf{8/10}& \textbf{3/10} & \textbf{36/50} \\
\midrule
$\pi_0$-FAST & \textbf{10/10} & \textbf{10/10} & 9/10 & 7/10 & 6/10   & 42/50 \\
$\pi_0$-FAST + \papername{}  & \textbf{10/10} & \textbf{10/10} & \textbf{10/10} & \textbf{9/10} & \textbf{8/10}  &  \textbf{47/50} \\
\midrule
$\pi_0$  & 7/10 &  7/10 & 4/10 & 5/10 & 4/10   & 27/50\\
$\pi_0$ + \papername{} & \textbf{10/10} & \textbf{ 10/10} & \textbf{7/10} & \textbf{8/10} & \textbf{8/10}   &  \textbf{43/50} \\
\addlinespace[1pt]
\hdashline
\addlinespace[1pt]
\multicolumn{2}{l}{\small \textit{Other baseline settings}} & &  & & & \\
$\pi_0$ + Traces~\cite{trace-vla} & 8/10  & 8/10 & 5/10 & 2/10 & 2/10 & 25/50\\
$\pi_0$ + RoboPoint~\cite{robopoint} & 8/10 & 9/10 & 4/10 & 6/10 & 0/10   & 27/50 \\
$\pi_0$ + Depth Images & 7/10  & 9/10 & 5/10 & 7/10 & 4/10  & 32/50  \\
\specialrule{0.3mm}{0mm}{0mm}
\end{tabular}
}
\caption{Performance comparison on five real-world tasks. Each task is evaluated over ten trials. 
}
\vspace{-10pt}
\label{tab:real-world}

\end{table*}

%% file: sections/5_conclusion.tex
\section{Conclusion and Discussions}


We present \papername{}, a lightweight visual augmentation technique that enhances visuomotor policy learning by embedding spatial cues, such as end-effector position, orientation, and grasp state, into RGB observations through scope reticles and shooting lines. 
This augmentation improves task performance by depicting spatial context directly in pixel space, offering grounded 2.5D visual guidance without requiring any changes to the model architecture. 
Experiments in both simulated and real-world environments demonstrate the effectiveness and generalizability of \papername{} across various vision-language-action backbones. 
In the future, we plan to enrich \papername{} with more semantic cues, such as object segmentation and affordance labels, and extend its applicability to diverse embodiments, including bi-manual and hand robots, to address more dexterous manipulation tasks.

%% file: sections/7_limitation.tex
\section*{Limitations}

While effective in our experimental setup using a Panda robot with a parallel-jaw gripper, \papername{} has several limitations:

\begin{itemize}[leftmargin=*]
\item \textbf{Dependence on depth sensing.} \papername{} relies on metric depth information that is aligned with RGB images. Consequently, it requires either an RGB-D sensor or a reliable monocular depth estimation model. Our method is highly efficient ($<$1ms) when sensor depth is available, while remaining compatible with model-predicted depth in RGB-only settings, but needs to afford  online depth model inference overhead.

\item \textbf{Assumption of nearby surfaces.} The visual guidance provided by \papername{}, such as shooting lines and reticles, assumes the end-effector is oriented toward a nearby or enclosed surface (e.g., tabletop tasks). In open-space scenarios, such as manipulation in mid-air or above empty space, these visual cues may project onto distant regions, reducing their utility as effective spatial indicators.

\item \textbf{Limited generalization to high-DOF effectors.} \papername{} is most suitable for simple end-effectors with limited degrees of freedom, such as parallel-jaw grippers. Extending the method to more dexterous end-effectors (e.g., anthropomorphic hands) would require complex designs to visually represent finger states and contact points, which may be nontrivial and reduce the method's generalizability to high-DOF manipulation tasks.

\item \textbf{Reduced effectiveness in constrained motion settings.} In tasks where the end-effector operates within a small spatial region and constantly holds an object (e.g., in-hand manipulation using tools), the projection length of the guidance cues becomes minimal. As a result, the encoded spatial information is limited and may provide little additional benefit to the visuomotor policy.


\end{itemize}

%% file: sections/6_appendix.tex
\newpage
\appendix
\section{Appendix}

\subsection{\papername{} Algorithm}
\label{sec: appx-alg}

\begin{algorithm}[H]
\caption{\textsc{Wld2Img}}
\begin{algorithmic}[1]
\Require 3D point $\mathbf{p}_{\text{wld}}$, extrinsic matrix $\mathbf{E} \in \mathbb{R}^{4 \times 4}$ and intrinsic matrix $\mathbf{K} \in \mathbb{R}^{3 \times 3}$ of camera $c$
\Ensure 2D image coordinates $(u_c, v_c)$ and depth $z_c$

\State Extract rotation $\mathbf{R} \gets \mathbf{E}[0{:}3,\ 0{:}3]$, translation $\mathbf{T} \gets \mathbf{E}[0{:}3,\ 3]$
\State $\mathbf{p}_\text{cam} \gets \mathbf{R} \cdot \mathbf{p}_{\text{wld}} + \mathbf{T}$ \Comment{Transform to camera frame}
\State $(x_c,\ y_c,\ z_c) \gets \mathbf{p}_\text{cam}$

\State Extract intrinsics: $f_x \gets \mathbf{K}_{00}$, $f_y \gets \mathbf{K}_{11}$, $c_x \gets \mathbf{K}_{02}$, $c_y \gets \mathbf{K}_{12}$
\State $u_c \gets \left\lfloor \frac{f_x x_c}{z_c} + c_x \right\rfloor$
\State $v_c \gets \left\lfloor \frac{f_y y_c}{z_c} + c_y \right\rfloor$

\State \Return $(u_c, v_c, z_c)$
\end{algorithmic}
\end{algorithm}

\begin{algorithm}[H]
\caption{\textsc{CheckVisibility}}
\begin{algorithmic}[1]
\Require Pixel coordinates $(u_c, v_c)$, depth $z_c$, depth map $D$, image size $(H, W)$, threshold $\epsilon$
\If{$z_c \leq 0$}
    \State \Return False \Comment{Behind the camera}
\ElsIf{$u \not\in [0, W)$ or $v \not\in [0, H)$}
    \State \Return False \Comment{Out of image bounds}
\EndIf
\If{$D[v_c, u_c] \leq z_c + \epsilon$}
    \State \Return False \Comment{Occluded by closer object}
\Else
    \State \Return True \Comment{Visible}
\EndIf
\end{algorithmic}
\end{algorithm}

\begin{algorithm}[H]
\caption{\textsc{FindStopPoint}}
\begin{algorithmic}[1]
\Require Starting position $\mathbf{p}$, orientation quaternion $q$, camera extrinsics $\mathbf{E}$, intrinsics $\mathbf{K}$, depth map $D$, image size $(H, W)$, step size $\delta$, tolerance number $N$
\State Convert $q$ to direction vector $\mathbf{d}$
\State Initialize empty lists $\text{points3D}$, $\text{points2D}$, $\text{visibility}$; Initialize tolerance count $n=0$
\While{step along $\mathbf{d}$ within 2 meter}
    \State $\mathbf{p} \gets \mathbf{p} + \delta \cdot \mathbf{d}$
    \State $(u, v, z) \gets$ \Call{Wld2Img}{$\mathbf{p}, \mathbf{E}, \mathbf{K}$}
    \State $vis \gets$ \Call{CheckVisibility}{$u, v, x, D, H, W$}
    \If{$n < N$ and $vis$ is False}
        \State $n \gets n+1$
    \ElsIf{$n >= N$}
        \State \textbf{break}
    \EndIf
    \State Append $\mathbf{p}$ to $\text{points3D}$, $(u, v)$ to $\text{points2D}$, $vis$ to $\text{visibility}$
\EndWhile
\State \Return $(\text{points3D}, \text{points2D}, \text{visibility})$
\end{algorithmic}
\end{algorithm}

\begin{algorithm}[H]
\caption{\textsc{RenderOnFixedCamera}}
\begin{algorithmic}[1]
\Require RGB image $I$, depth $D$, extrinsics $\mathbf{E}$, intrinsics $\mathbf{K}$, gripper state $(\mathbf{p}, q, open)$, image size $(H, W)$
\State $(u, v, z) \gets$ \Call{Wld2Img}{$\mathbf{p}, \mathbf{E}, \mathbf{K}$}
\If{\Call{CheckVisibility}{$u, v, z, D, H, W$} is not visible}
    \State \Return $I$
\EndIf
\State $(\_,  \text{points2D},  \text{visibility}) \gets$ \Call{FindStopPoint}{$\mathbf{p}, q, \mathbf{E}, \mathbf{K}, D, H, W$}
\If{$open$}
    \State Draw \textcolor{lightgreen}{green} line on $I$ over longest visible span using  \text{points2D}
\Else
    \State Draw \textcolor{purple}{purple} line on $I$ through all visible spans using  \text{points2D}
\EndIf
\State Draw center dot at $(u, v)$ with color based on $open$
\State \Return $I$
\end{algorithmic}
\end{algorithm}

\begin{algorithm}[H]
\caption{\textsc{RenderOnWristCamera}}
\begin{algorithmic}[1]
\Require Wrist RGB $I_w$, depth $D_w$, extrinsics $\mathbf{E}_w$, intrinsics $\mathbf{K}_w$, gripper state $(\mathbf{p}, q, open)$, image size $(H, W)$, MaxEEtoSurfaceDistance, MinReticleLength, MaxReticleLength
\State $(u_w, v_w, z_w) \gets$ \Call{Wld2Img}{$\mathbf{p}, \mathbf{E}_w, \mathbf{K}_w$}
\If{\Call{CheckVisibility}{$u_w, v_w, z_w, D_w, H, W$} is not visible}
    \State \Return $I_w$
\EndIf
\State $(\_,  \text{points2D},  \text{visibility}) \gets$ \Call{FindStopPoint}{$\mathbf{p}, q, \mathbf{E}_w, \mathbf{K}_w, D_w, H, W$}
\State Compute \textit{center} $\gets$ last visible point in  \text{points2D} or $(u_w, v_w)$

\State Compute \textit{dynamic line length} with:
\[
 \textit{scaling} \gets \max\left( \frac{\text{MaxEEtoSurfaceDistance} - z_w}{\text{MaxEEtoSurfaceDistance}},\ 0 \right)\]
\[
\textit{line\_length} \gets \text{MinReticleLength} + \textit{scaling} \times (\text{MaxReticleLength} - \text{MinReticleLength})
\]

\State Add crosshair centered at \textit{center} with \textit{line\_length}, adapting color based on $open$
\State \Return $I_w$

\end{algorithmic}
\end{algorithm}

\newpage
\subsection{Ablation Settings}
\label{sec: appx-ablation}
To further investigate how different designs of \papername{} reticles affect task performance, we conduct a comprehensive comparison of several visual cue variants. 
Specifically, we evaluate $\pi_0$ and $\pi_0$-FAST under the following settings:
\begin{enumerate}
    \item \textsc{w/ plain color}: using a uniform gray color for the augmented guidance;
    \item \textsc{w/ grasp sense}: detecting whether an object is in between the gripper fingers and changing the reticle color to indicate a successful grasp;
    \item  \textsc{w/ fixed length}: rendering crosshair reticles with a fixed line length instead of adapting to the estimated depth;
    \item \textsc{w/ small scale}: reducing the size and thickness of the visual cues;
    \item \textsc{w/ bullseye style}: replacing the crosshair with a classical bullseye (concentric circle) reticle design;
\end{enumerate}

\begin{figure}
    \centering
    \includegraphics[width=0.98\linewidth]{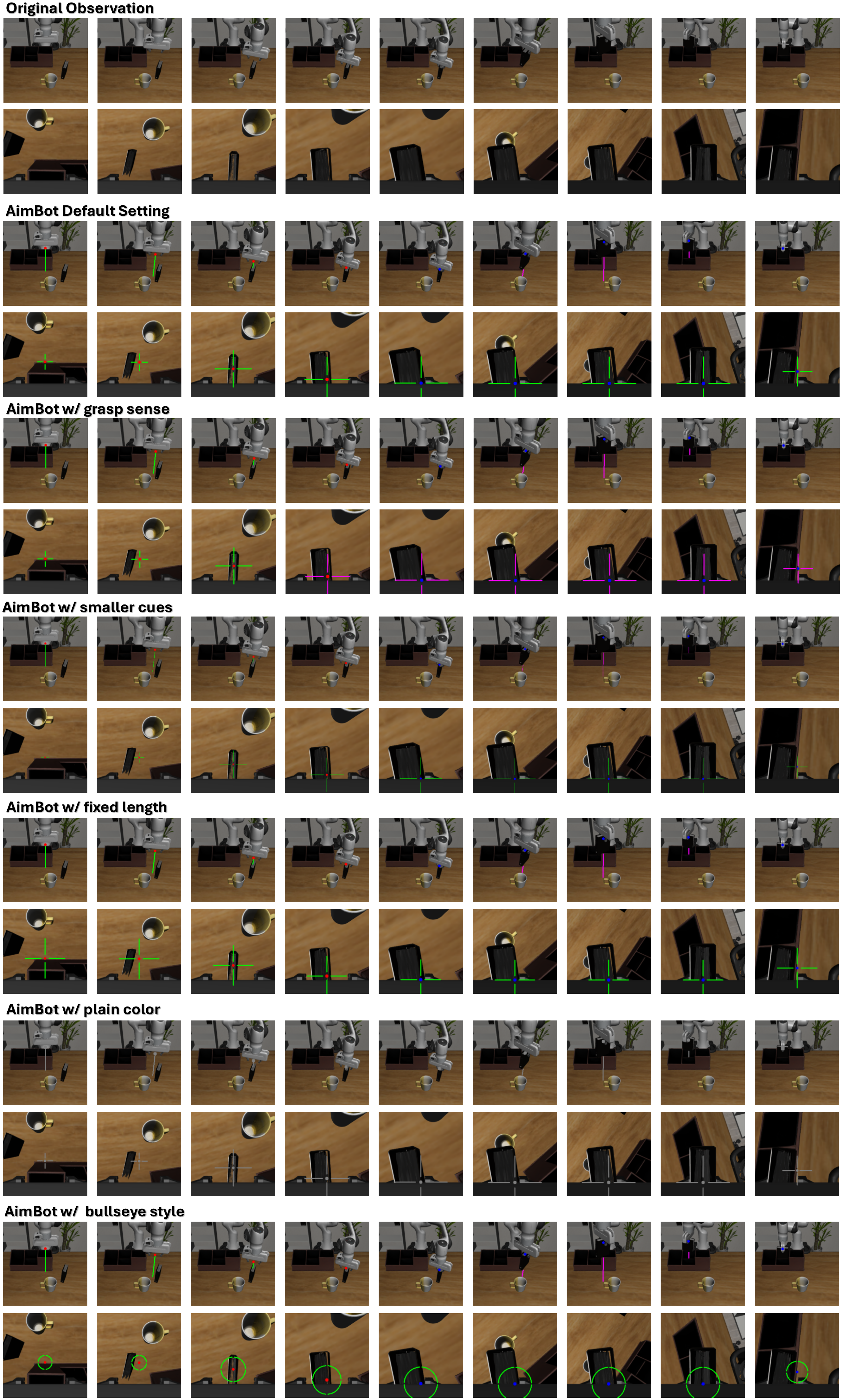}
    \caption{Comparison of different \papername{} variants.}
    \label{fig: ablation}
\end{figure}

Figure \ref{fig: ablation} illustrates different ablated settings of our \papername{} augmentation. Results of $\pi_0$-FAST and $\pi_0$ are given in Table \ref{tab:libero-ablation1} and Table \ref{tab:libero-ablation2}, respectively. As we can see, the default setting achieve the best performance in average. Interestingly, even with plain color,  the model can still achieve adequate performance, indicating the importance of visual cues for spatial information. 

\begin{table}[htp]
\centering
\scalebox{0.88}{
\begin{tabular}{lccccc}
\toprule
\papername{} Setting & 
\begin{tabular}[c]{@{}c@{}} \texttt{LIBERO} \\ \texttt{Spatial} \end{tabular}
& \begin{tabular}[c]{@{}c@{}} \texttt{LIBERO} \\ \texttt{Object} \end{tabular} & \begin{tabular}[c]{@{}c@{}} \texttt{LIBERO} \\ \texttt{Goal} \end{tabular} & \begin{tabular}[c]{@{}c@{}} \texttt{LIBERO} \\ \texttt{Long} \end{tabular} & \begin{tabular}[c]{@{}c@{}} Avg. \\ Success \end{tabular} \\
\midrule
Default  & 97.2 & 96.5 & \textbf{94.0} & \textbf{87.1} & \textbf{93.7}\\
\quad \textsc{w/ plain color} & \textbf{97.4} & 97.6 & 93.2 & 84.0 & 93.1 \\
\quad  \textsc{w/ grasp sense} & 95.8 & 97.0 & 93.6 & 81.6 & 92.0 \\
\quad  \textsc{w/ fixed length} & 95.8 & \textbf{97.8} & 90.8 & 84.6 & 92.3 \\
\quad  \textsc{w/ small scale} & 94.6 & \textbf{97.8} & 89.6 & 85.2 & 91.8 \\
\quad \textsc{w/ bullseye style} & 94.8 & 96.8 & 91.0 & 82.4 & 91.3 \\
\bottomrule
\end{tabular}
}
\vspace{5pt}
\caption{Ablation Study for $\pi_0$-FAST for different \papername{} settings.}
\label{tab:libero-ablation1}
\end{table}

\begin{table}[htp]
\centering
\scalebox{0.88}{
\begin{tabular}{lccccc}
\toprule
\papername{} Setting & 
\begin{tabular}[c]{@{}c@{}} \texttt{LIBERO} \\ \texttt{Spatial} \end{tabular}
& \begin{tabular}[c]{@{}c@{}} \texttt{LIBERO} \\ \texttt{Object} \end{tabular} & \begin{tabular}[c]{@{}c@{}} \texttt{LIBERO} \\ \texttt{Goal} \end{tabular} & \begin{tabular}[c]{@{}c@{}} \texttt{LIBERO} \\ \texttt{Long} \end{tabular} & \begin{tabular}[c]{@{}c@{}} Avg. \\ Success \end{tabular} \\
\midrule
Default  & \textbf{96.9} & \textbf{98.4} & \textbf{97.2} & \textbf{91.0} & \textbf{95.9}\\
\quad \textsc{w/ plain color} & 96.8 & 97.2 & 96.8 & 89.0 & 95.0 \\
\quad  \textsc{w/ grasp sense} & 95.0 & 98.0 & 95.4 & 86.8 & 93.8  \\
\quad  \textsc{w/ fixed length} & 96.0 & 96.2 & 94.6 & 87.2 & 93.5 \\
\quad  \textsc{w/ small scale} & 95.0 & 98.0 & 95.4 & 86.8 & 93.8 \\
\quad \textsc{w/ bullseye style} & 97.2 & 96.2 & 93.2 & 86.4 & 93.3 \\
\bottomrule
\end{tabular}
}
\vspace{5pt}
\caption{Ablation Study for $\pi_0$ for different \papername{} settings.}
\label{tab:libero-ablation2}
\end{table}

\newpage
\subsection{\papername{} Real World Set-up and Dataset Statistics}
\label{appendix:dataset}
\begin{figure}[hp]
    \centering
    \includegraphics[width=0.98\linewidth]{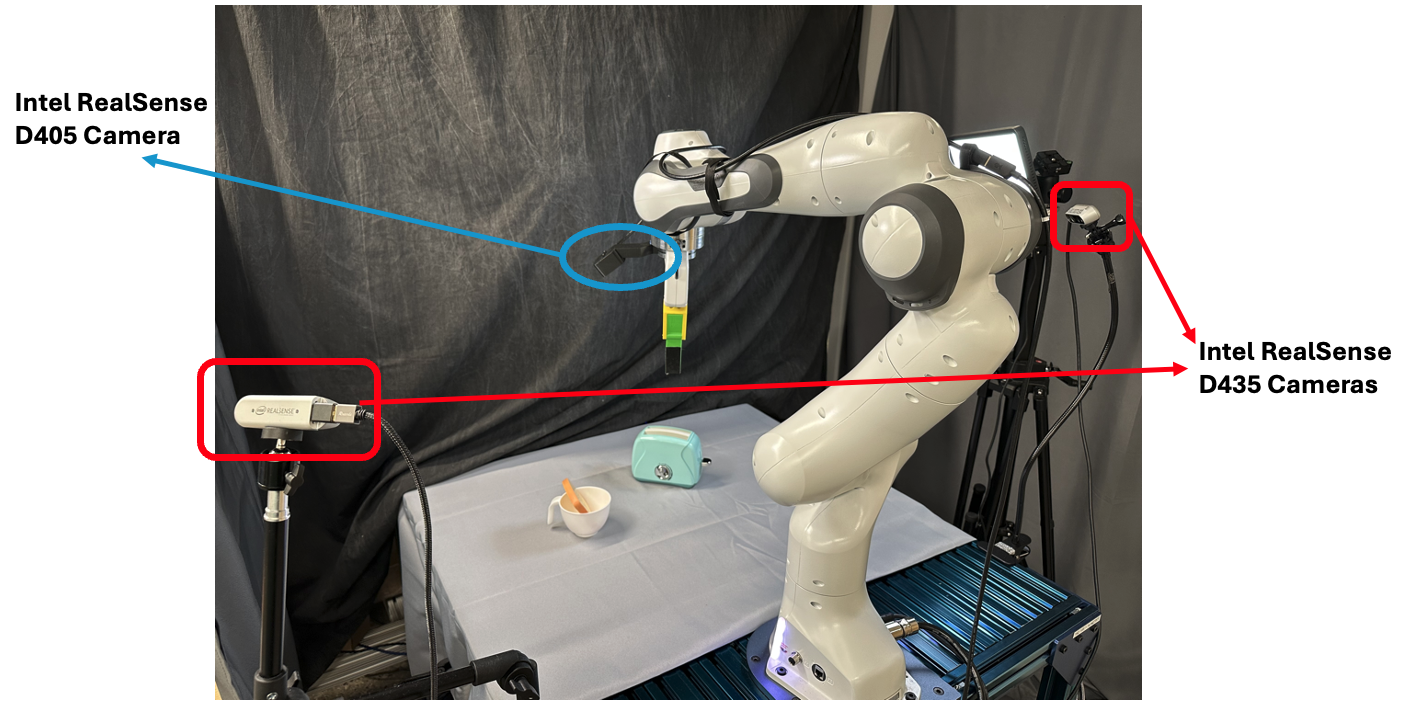}
    \caption{Our real robot platform setup.}
    \label{fig:robot-set-up}
\end{figure}

Figure~\ref{fig:robot-set-up} shows the Franka Emika Panda robot setup used in our real-world experiments.
Table~\ref{tab:realworld-dataset-stats} provides detailed statistics of the collected dataset, including language goals, number of demonstrations, required skills, and average episode length in environment timesteps.
Additionally, Figure~\ref{fig:real_world_rollouts_detail} presents sample rollouts for each real-world task, showcasing the use of \papername{} visual guidance.

\input{tables/appendix_real_world_data_stats}

\input{floating/real_world_rollouts}

\clearpage
\newpage
\subsection{Visualization of attention maps}
\label{appx: attention}
\begin{figure}[h]
    \centering
    \includegraphics[width=1.0\linewidth]{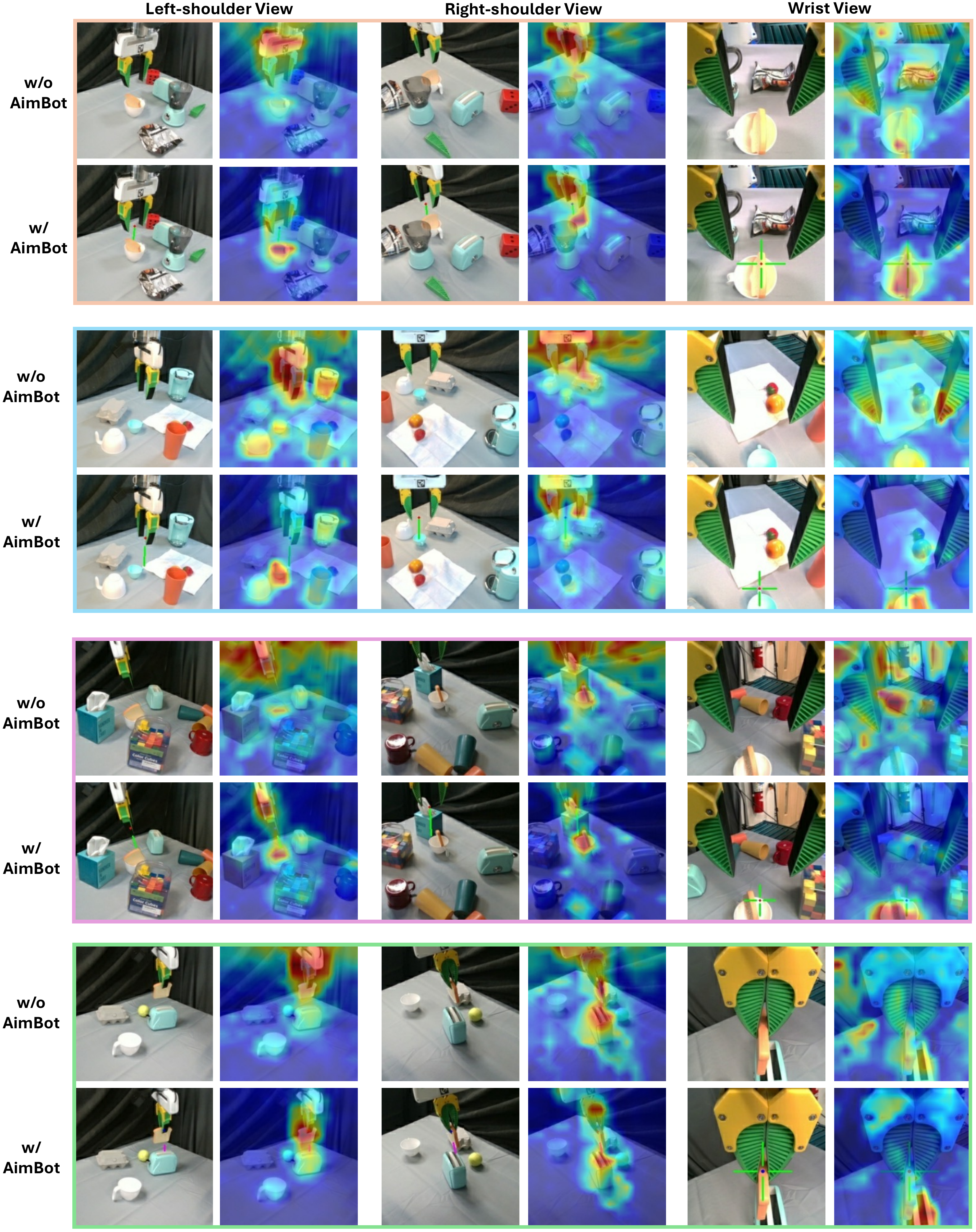}
    \caption{Additional attention heatmap examples comparing w/ and w/o \papername{}.}
    \label{fig:more_vis}
\end{figure}

Figure~\ref{fig:more_vis} shows more examples of attention visualizations from the OpenVLA-OFT model, trained with and without our proposed \papername{}. The visualized attention maps correspond to the 1st layer 11th head in the LLM backbone, summing up all attention probabilities from the action prediction tokens to the image patches. To generate the heatmaps, we project the patch-wise attention scores onto the original image using an upsampling method as in GradCAM\cite{jacobgilpytorchcam}.
We observe that incorporating \papername{} leads to more focused attention on task-relevant objects, thereby enhancing the model's spatial understanding and improving manipulation performance.

\newpage
\subsection{Misalignment Failure Examples}
\label{sec: appx-failure}
\begin{figure}[h]
    \centering
    \includegraphics[width=0.9\linewidth]{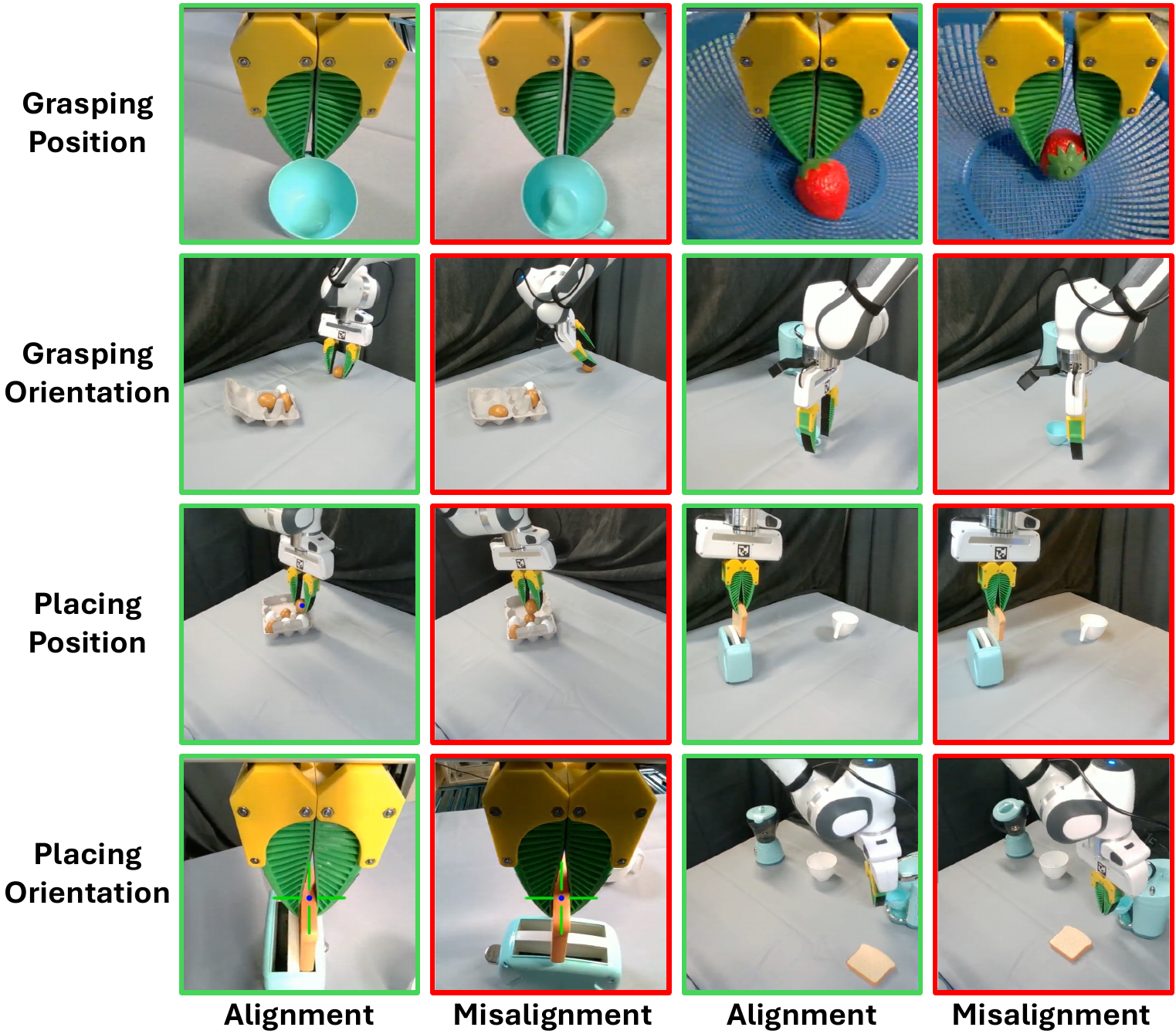}
    \caption{Examples of four misalignment types in real-world robot trials: grasping position, grasping orientation, placing position, and placing orientation. Each row shows representative wrist-view or shoulder-view images. For each misalignment type, we include aligned  (\textcolor{lightgreen}{green} borders) and misaligned (\textcolor{red}{red} borders) examples to contrast correct behavior with failure cases.}
    \label{fig:failure}
\end{figure}

We categorize all failure trials into different types: (1) grasping position misalignment, (2) grasping orientation misalignment, (3) placing position misalignment, (4) placing orientation misalignment, and (5) other non-misalignment failures (e.g., getting stuck, failed non-prehensile skills, or performing actions in the wrong task order). If a failure trial involves multiple misalignment cases, we attribute the failure to the last observed misalignment. Examples of such misalignments that led to failures in our real-world experiments are illustrated in Figure~\ref{fig:failure}.

\newpage
\subsection{Comparison of Proprioceptive State Representation}
\label{sec: appx-prior}
We compare different ablation of $\pi_0$ variants training with or without \papername{} and using or not using proprioceptive state vector encoding. The results on the LIBERO benchmark are given in Table \ref{tab: pi0-variant}.

\begin{table}[h]
    \centering
    \scalebox{0.9}{
    \begin{tabular}{lccccc}
    \toprule
    \textbf{Model} & 
    \begin{tabular}[c]{@{}c@{}} \texttt{LIBERO} \\ \texttt{Spatial} \end{tabular}
    & \begin{tabular}[c]{@{}c@{}} \texttt{LIBERO} \\ \texttt{Object} \end{tabular} & \begin{tabular}[c]{@{}c@{}} \texttt{LIBERO} \\ \texttt{Goal} \end{tabular} & \begin{tabular}[c]{@{}c@{}} \texttt{LIBERO} \\ \texttt{Long} \end{tabular} & \begin{tabular}[c]{@{}c@{}} \textsc{Average} \\ \textsc{Success Rate} \end{tabular} \\
    \midrule
    $\pi_0$ $+$ \papername{} & \textbf{96.9}	&98.4	& \textbf{97.2} & \textbf{91.0} & \textbf{95.9}\\
    $\pi_0$ $+$ \papername{} $-$ proprio. & 96.6	& 96.8	& 94.8 & 88.0 & 94.1\\
    $\pi_0$  & 96.8	& \textbf{98.8}	& 95.8 & 85.2 & 94.2 \\
    $\pi_0$ $-$ proprio. & 96.6	& 96.4	& 94.8 & 83.2 & 92.8 \\
    $\pi_0$ $+$ \papername{} (random) & 93.4	&92.0	&89.8 & 77.4 & 88.1\\ 
    \bottomrule
    \end{tabular}}
\vspace{3pt}
\caption{Comparison of different proprioceptive state representations.}
\label{tab: pi0-variant}
\end{table}

\subsection{Out-of-distribution Scenes}
\label{sec: appx-ood}
Out-of-distribution (OOD) examples are illustrated in Figure~\ref{fig:ood_scenes}.
All OOD evaluation rollouts can be found at \url{https://aimbot-reticle.github.io/}.

\input{floating/ood_scenes}


%% file: tables/appendix_real_world_data_stats.tex
\begin{table}[htp]
\centering
\scalebox{0.88}{
\begin{tabular}{clccc}
\toprule
\textbf{Tasks} & \textbf{Language Goal} & \textbf{Demos} & \textbf{Skills} & \textbf{Mean Ep. Length} \\
\midrule
\begin{tabular}[c]{@{}c@{}} \texttt{Fruits} \\ \texttt{in Box} \end{tabular} & Put all the fruits into the box. & 77 & \begin{tabular}[c]{@{}c@{}} Pick. Place. \end{tabular} & 366 \\
\addlinespace[2pt]
\hdashline
\addlinespace[2pt]
\begin{tabular}[c]{@{}c@{}} \texttt{Ball in} \\ \texttt{Drawer} \end{tabular} & Put the tennis ball inside the drawer. & 80 & \begin{tabular}[c]{@{}c@{}} Open. Grasp. \\ Place. Close. \end{tabular} & 390 \\
\addlinespace[2pt]
\hdashline
\addlinespace[2pt]
\begin{tabular}[c]{@{}c@{}} \texttt{Bread in} \\ \texttt{Toaster} \end{tabular} & Put the bread inside the toaster. & 120 & \begin{tabular}[c]{@{}c@{}} Grasp. Reorient. \\ Insert. \end{tabular} & 251 \\
\addlinespace[2pt]
\hdashline
\addlinespace[2pt]
\begin{tabular}[c]{@{}c@{}} \texttt{Place} \\ \texttt{Coffee Cup} \end{tabular} & Put the cup on the coffee machine. & 120 & \begin{tabular}[c]{@{}c@{}} Grasp. Reorient. \\ Place. \end{tabular} & 270 \\
\addlinespace[2pt]
\hdashline
\addlinespace[2pt]
\begin{tabular}[c]{@{}c@{}} \texttt{Egg in} \\ \texttt{Carton} \end{tabular} & Put the eggs inside the egg carton. & 151 & \begin{tabular}[c]{@{}c@{}} Grasp. Reorient. \\ Place. Close. \end{tabular} & 319 \\
\bottomrule
\end{tabular}
}
\vspace{5pt}
\caption{Real-world dataset statistics for training multi-task policies. Mean episode lengths are given in environment timesteps.}
\label{tab:realworld-dataset-stats}
\end{table}

%% file: floating/real_world_rollouts.tex
\begin{figure}[h]
    \centering
    \includegraphics[width=1.0\textwidth, keepaspectratio]{pdfs/real_world_rollouts_v8.pdf}
    \vspace{-10pt}
    \caption{\textbf{Task Examples and Visualization of \papername{}'s in Multi-View Observations.} We test our method on five challenging tasks: \textbf{\texttt{Fruits in Box}}, \textbf{\texttt{Tennis Ball in drawer}}, \textbf{\texttt{Bread in Toaster}}, \textbf{\texttt{Place Coffee Cup}} and \textbf{\texttt{Egg in Carton}}. Here we show how our \papername{} augment the shooting line and scope reticle in the left shoulder and wrist camera views, respectively.}
    \vspace{-5pt}
    \label{fig:real_world_rollouts_detail}
\end{figure}


%% file: floating/ood_scenes.tex
\begin{figure}[h]
    \centering
    \includegraphics[width=\textwidth]{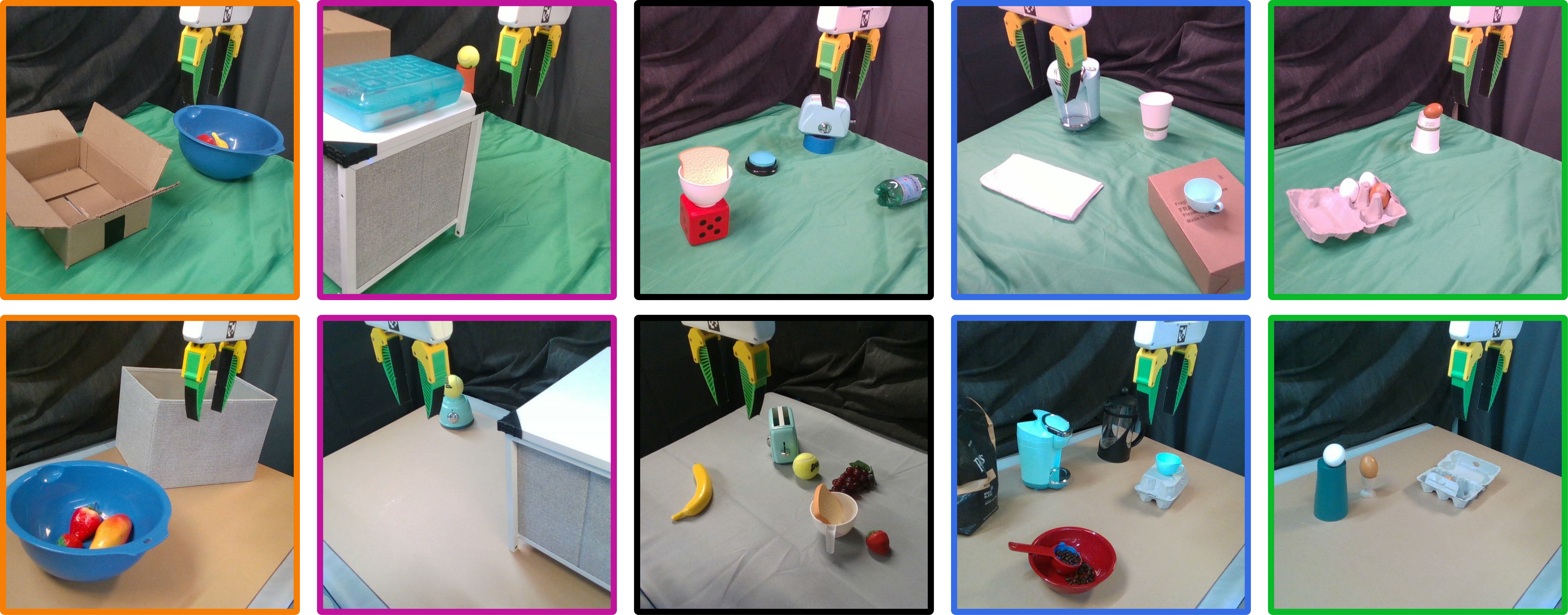}
    \caption{Sample out-of-distribution evaluation scenes to test generalization of policies to different backgrounds, varying lighting conditions (e.g. flashing lights, cool/warm lights), unseen distractors with real-time human perturbations, and varying camera poses, where \papername{} bridges the distribution gap by grounding to useful depth information instead of plain visual features.}
    \label{fig:ood_scenes}
\end{figure}